\documentclass{article}

\usepackage{arxiv}

\usepackage[utf8]{inputenc} 
\usepackage[T1]{fontenc}    
\usepackage{hyperref}       
\usepackage{url}            
\usepackage{booktabs}       
\usepackage{amsfonts}       
\usepackage{nicefrac}       
\usepackage{microtype}      
\usepackage{lipsum}

\usepackage{amsmath}
\usepackage{amssymb}
\usepackage{multirow}
\usepackage{subfigure,graphicx}
\usepackage{courier}
\usepackage{booktabs}
\usepackage{latexsym,amsmath,amssymb,graphicx}
\usepackage{url}
\usepackage{color}
\usepackage{bm}
\usepackage{helvet}
\usepackage{algorithm}
\usepackage{algorithmic}

\title{LAConv: Local Adaptive Convolution for Image Fusion}

\author{
	Zi-Rong Jin\\
	School of Optoelectronic Science and Engineering\\
	University of Electronic Science and Technology of China\\
	Chengdu, 611731, China\\
	\texttt{2018051403016@std.uestc.edu.cn} 
	\And
  Liang-Jian Deng\protect{\footnote{Corresponding author}}\\
  School of Mathematical Sciences\\
  University of Electronic Science and Technology of China\\
  Chengdu, 611731, China\\
  \texttt{liangjian.deng@uestc.edu.cn}   
\And
   Tai-Xiang Jiang\\
   School of Economic Information Engineering\\
   Southwestern University of Finance and Economics\\
   Chengdu, China\\
   \texttt{taixiangjiang@gmail.com} \And
   Tian-Jing Zhang\\
   Yingcai Honors College\\
   University of Electronic Science and Technology of China\\
   Chengdu, 611731, China\\
   \texttt{zhangtianjinguestc@163.com} \\
}

\begin{document}
\maketitle

\begin{abstract}
	The convolution operation is a powerful tool for feature extraction and plays a prominent role in the field of computer vision. However, when targeting the pixel-wise tasks like image fusion, it would not fully perceive the particularity of each pixel in the image if the uniform convolution kernel is used on different patches. In this paper, we propose a local adaptive convolution (LAConv), which is dynamically adjusted to different spatial locations. LAConv enables the network to pay attention to every specific local area in the learning process. Besides, the dynamic bias (DYB) is introduced to provide more possibilities for the depiction of features and make the network more flexible. We further design a residual structure network equipped with the proposed LAConv and DYB modules, and apply it to two image fusion tasks. Experiments for pansharpening and hyperspectral image super-resolution (HISR) demonstrate the superiority of our method over other state-of-the-art methods. It is worth mentioning that LAConv can also be competent for other super-resolution tasks with less computation effort.
\end{abstract}


\section{Introduction}
Modeling the relationship between pixel and pixel, patch and patch, even image and image is a fundamental problem in computer vision, where applications are ranging
from single image super-resolution, image restoration, to image fusion. Convolution is indeed an effective tool to solve this problem. By performing convolution operations on the image, specific important features can be extracted. And through the stacked convolutional layers, learning ability of the network can be strengthened, thus end-to-end relational mapping can be realized. However, when faced with refined pixel-wise tasks, standard convolution operations are often unable to focus on each pixel of the image, and the accuracy can only be improved by deepening the network, resulting in a cumbersome model.

\begin{figure}[!htp]
	\begin{center}
		{\includegraphics[width=0.6\linewidth]{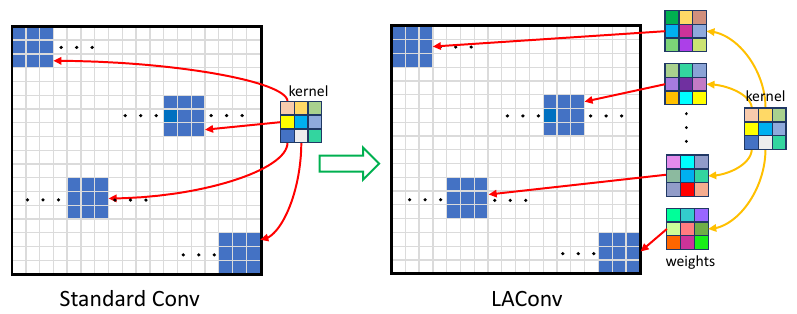}}
	\end{center}
	\caption{{A toy example of the motivation in this paper. Left: The standard convolution, by which all patches in the feature map are convolved by a uniform kernel; Right: The proposed local adaptive convolution (LAConv), by which all patches in the feature map are convolved by local adaptive kernels obtained by a uniform kernel multiplied with weight. The red line means the convolutional operation, and the orange line is the pixel-wise multiplication between the kernels and the weights generated from the local patch.}}\label{Fig:toy}
\end{figure}
This paper focuses on image fusion, which aims to improve the spatial resolution and geometric accuracy of images through appropriate fusion strategies. It is a task that has a wide range of vital applications including the following two: 1) Remote sensing image pansharpening, which fuses low-resolution multispectral image (LR-MSI) and high-resolution panchromatic image (HR-PANI), makes up for the deficiencies of a certain kind of remote sensing data, and promotes the applicability of remote sensing image. 2) Hyperspectral image super-resolution (HISR), which fuses a low-resolution hyperspectral image (LR-HSI) and a high-resolution multispectral image (HR-MSI), takes advantage of different types of images, finally obtains a high-resolution hyperspectral image (HR-HSI). In this work, we mainly address these two fusion tasks.

\begin{figure*}[!htp]
	\begin{center}
		{\includegraphics[width=1\linewidth]{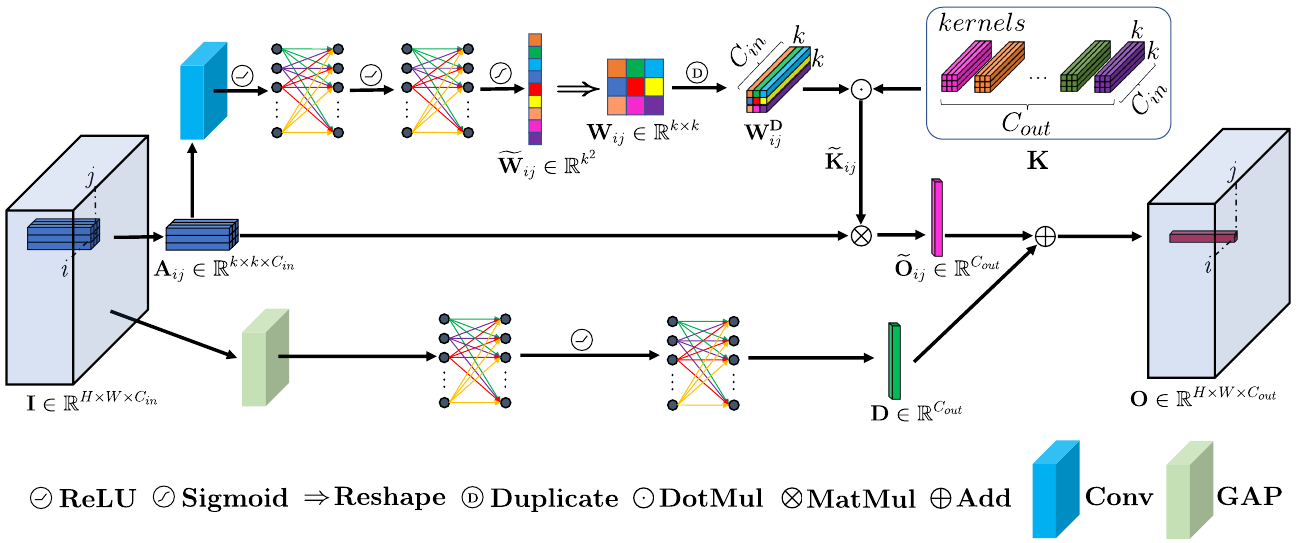}}
	\end{center}
	\caption{{The interpretation of LAConv consists of two parts. The upper part is the local adaptive convolution (LAConv), and the bottom part is the dynamic bias (DYB) strategy. For better understanding, we take kernel size $k=3$ as an expanded demonstration. $\odot$ represents the pixel-wise multiplication (denoted as DotMul) of $\mathbf{W^{D}}_{ij}$ and every kernel in $\mathbf{K}$}.}\label{Fig:laconv}
\end{figure*}

Whether it is pansharpening or HISR, the difficulty lies in achieving a balance between spatial resolution and spectral resolution. An ideal solution is that we can pay attention to pixel-level information, and perform feature representation with pixel uniqueness while ensuring that the global information is not distorted. It is undeniable that the method based on convolutional neural networks (CNNs) has significantly promoted the development of image fusion technology. However, most of the current methods that emerge in an endless stream are working on changing network structure and depth, and the performance of the network mainly depends on the number of network parameters.

In addition, the existing CNN-based methods for image fusion have several fundamental problems. Firstly, for an image, its low-frequency components occupy the main part. In contrast, the high-frequency components that represent texture account for a much smaller proportion. Therefore, in order to minimize the loss, the network will concentrate on most of the low-frequency patches during the learning process. In other words, the convolution kernels will update the parameters following the direction that is conducive to the super-resolution of the low-frequency patches. Instead, the features of the high-frequency patches are ignored, and further leads to the undesirable smoothness of the fusion results. However, the evaluation of image quality is highly dependent on these high-frequency components. Secondly, in standard convolution, the update and optimization of the bias are all based on the average value of the pixels in the feature map. And the bias of one feature map for different samples are fixed, which restricts the network flexibility. Therefore, for the fusion tasks, the local characteristics are exactly what needs to be paid attention to, and the conventional bias can also be improved. The key to the tasks is the mutual coordination and supplementation of high-frequency information in spatial and spectral dimensions and the repair of local details.

In this paper, we propose a novel local adaptive convolution (LAConv), in which each single  kernel will be scaled with a adaptive weight generated from the local patch. Besides, dynamic bias (DYB), which yields from the whole given feature map, is adopted to provide the global information. To a certain extent, DYB can be seen as a novel channel attention mechanism. With the combination of the weighted convolution kernel and DYB, the LAConv has therefore become more flexible and has more powerful capability of local feature representation. In addition, we embed LAConv into a residual structure to construct a simple network, which can achieve satisfactory results with a small computational cost. Extensive experiments indicate the remarkable effectiveness and efficiency of LAConv in image fusion tasks.

In brief, the contributions of this work are as follows:
\begin{itemize}
	\item We propose a new local adaptive convolution (LAConv), which generates an adaptive kernel based on each pixel and its neighbors. LAConv not only inherits all the advantages of standard convolution, but also enhances the ability of focusing on local features.
	\item A dynamic bias (DYB) strategy is introduced in LAConv. Compared with the conventional bias in the standard convolution, the DYB can supplement the global information into the local features, thus mitigate the subtle distortion caused by spatial discontinuities, further making the network more flexible.
	\item A simple residual network based on LAConv (LAResNet) is designed and applied to two image fusion tasks. Experiments demonstrate that benefit from proposed low-cost and easy-to-implement LAConv, LAResNet can achieve surpassing performance over the state-of-the-art methods even if it does not have deep layers and huge parameters.
	
\end{itemize}

\begin{figure*}[!htp]
	\begin{center}
		{\includegraphics[width=1\linewidth]{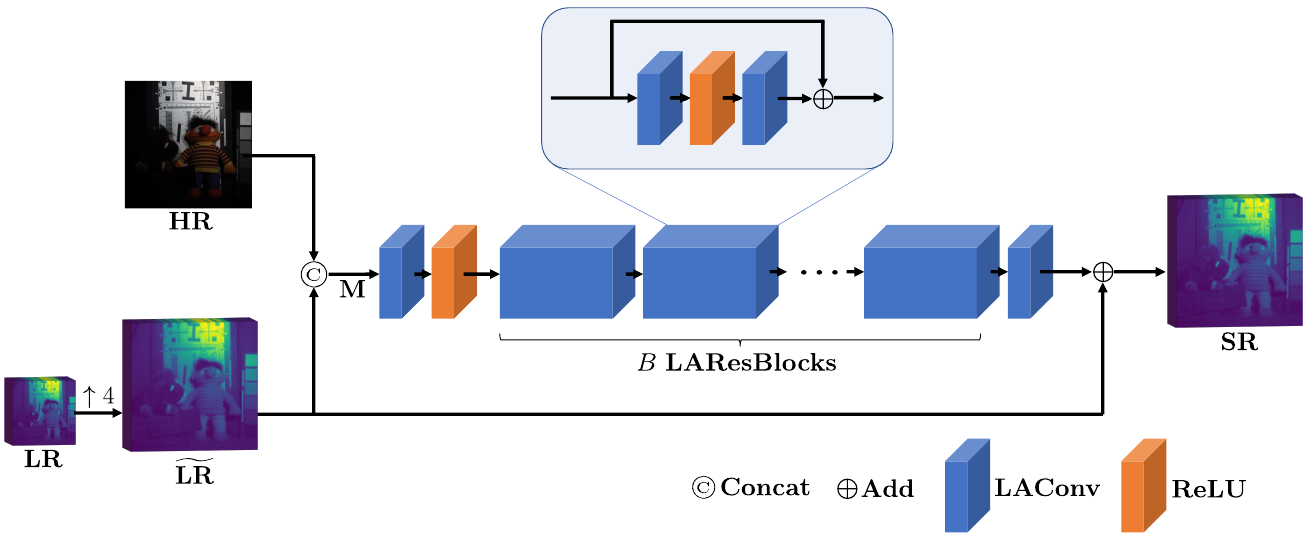}}
	\end{center}
	\caption{{Overall architecture of the proposed LAResNet. It takes HISR as example here (the image is from CAVE dataset). We can replace the ``LR'' and ``HR'' by other inputs for different fusion tasks, e.g., pansharpening.}}\label{Fig:laresnet}
\end{figure*}

\section{Related Works}

In this section, we first present the distinction between our work and other dynamic convolution methods for other applications and then introduce a series of previous advanced works on pansharpening and HISR. Finally, our motivation is stated.

\subsection{Dynamic Convolution}

In order to improve the performance of the model, the existing standard convolution neural network can mainly increase the parameters, depth, and number of channels, resulting in the model being too complex. Many pioneers have aware of this bottleneck. In~\cite{condconv}, Yang {\it et al.} proposed conditionally parameterized convolutions (CondConv), which breaks the traditional static convolution characteristics by calculating the convolution kernel parameters through the input samples, which is effective in inference. Different from our LAConv, the kernel dynamically generated by CondConv is for the entire sample, not for specific regions or pixels. Another notable work is dynamic convolution (DYConv) proposed in~\cite{dynet}, it aggregates multiple convolution kernels according to their customized attention degree to each input sample. Compared with standard convolution, it significantly improves the representation ability of the network but increases much computational parameters. In addition, it still focuses on the entire sample rather than the specific regions or pixels. Recently, a dynamic region-aware convolution (DRConv) was proposed in~\cite{drconv}, which automatically allocates multiple filters to spatial regions with similar characteristics, achieves satisfactory performance on many tasks ({\it i.e.}, classification, segmentation, and face recognition). The same idea as DRConv is that the proposed LAConv also has the characteristics of translation-invariance, but in contrast, LAConv only assigns weights to the convolution kernel and more completely retains the core of the standard convolution. Another difference between DRConv and ours is that DRConv has one more step to classify the region of the input feature map, this step is redundant in the fusion task. 

In addition to the above mentioned, an important point is that the above three dynamic convolutions all include the operation of global average pooling (GAP). We hold the opinion that GAP will cause information distortion in spatial dimension thus it is not suitable for image fusion tasks.

\subsection{Pansharpening}
\label{pansharpening}
Pansharpening is a challenging task in the field of remote sensing. The existing methods can be divided into traditional methods and deep learning (DL) methods based on big data-driven. Some classic traditional methods are the smoothing filter-based intensity modulation (SFIM)~\cite{sfim}, the generalized Laplacian pyramid (GLP)~\cite{glp} with MTF-matched filter~\cite{mtf} and regression-based injection model (GLP-CBD)~\cite{glp_cbd}, and the band-dependent spatial-detail with local parameter estimation (BDSD)~\cite{bdsd}.

Numerous DL-methods based on CNN have emerged recently, pushing the task of pansharpening to a new era, alleviating the distortions more or less existed in traditional methods. Typical works are PanNet~\cite{pannet}, DiCNN1~\cite{dicnn}, DMDNet~\cite{dmdnet}, and FusionNet~\cite{fusionnet}. What they have in common is the use of the same convolution kernel and conventional bias for feature extraction, resulting in limited learning capabilities of the network.

\subsection{HISR}
\label{hisr}
Similarly, previous work for HISR can be classified as traditional methods and DL-methods. The SOTA traditional methods including FUSE~\cite{FUSE}, the coupled sparse tensor factorization (CSTF)~\cite{CSTF} method and the cnn denoiser (CNN-FUSE)~\cite{CNN-FUS}. And advanced DL-methods including SSRNet~\cite{SSRNET}, ResTFNet~\cite{ResTFnet}, and MHFNet~\cite{MHFNet}. They all showed exceptional performance, but HISR also requires separate modeling of the characteristics of each pixel in order to better achieve super-resolution.

\subsection{Motivation}
The essence of the image fusion problem lies in the mutual complementation of information and the enhancement of spatial and spectral resolution. After in-depth consideration, we hold that the uniform convolution kernel and the conventional bias are not adequate for these pixel-to-pixel tasks. Specifically, the key high-frequency information can not be given special treatment, even be ignored during the learning process, leading to an undesirable smoothness of results in the super-resolution tasks. And the addition of a fixed value and the overall feature map is probably meaningless in this task. In order to address these issues, we propose LAConv and DYB, which we will introduce in detail in the following section.

\section{Methods}
\label{Method}
In this section, the procedure of LAConv and DYB is detailed firstly. After that, the structure of the proposed LAResNet will be expressed. 

\subsection{Local Adaptive Convolution}
Image fusion needs to accurately determine the value of each pixel, the specific situation is restoration and reconstruction of a pixel is only related to its neighbors, and it is weakly related to pixels that are far away from it. In order to fully explore the local information of the pixel, we have made a change in the design of the convolution kernel. While retaining the original convolution kernel, we have updated the state of the convolution kernel for each pixel. The specific operation is described below.

\textbf{Standard Convolution} Firstly, let us review the standard convolution. Consider a standard convolution without bias operates on a pixel $\textbf{I}_{ij} \in \mathbb{R}^{1\times 1\times C_{in}}$ that located at spatial coordinates $(i, j)$, its local patch is defined as $\textbf{A}_{ij} \in \mathbb{R}^{k\times k\times C_{in}}$, where $C_{in}$ and $k$ mean the channels of the inputting feature map and the patch size, respectively. During the standard convolution operation, all the local patches of the inputting feature map use the same kernels $\mathbf{K}$. Thus the operation can be expressed as follows:
\begin{equation}\label{eq:3}
\mathbf{\widehat{O}}_{ij} = \mathbf{A}_{ij}\otimes \mathbf{K} ,
\end{equation}
where $\textbf{K} \in \mathbb{R}^{C_{in} \times k\times k\times C_{out}}$ can be viewed as $C_{out}$ convolution kernels with the size $k\times k\times C_{in}$ on one layer. $\otimes$ represents the convolution operation in conventional CNN (also can be viewed as the operation of matrix multiplication (MatMul)). $\mathbf{\widehat{O}}_{ij} \in \mathbb{R}^{1\times 1\times C_{out}}$ is the result after the convolution, where $C_{out}$ mean the channels of the outputting feature map.

\textbf{LAConv} Different from the standard convolution, the kernel in our LAConv is automatically adjusting depending on the local patch. Let $\mathbf{\widetilde{K}}_{ij}\in \mathbb{R}^{C_{in} \times k\times k\times C_{out}}$ represents the kernel which is used to perform the convolution on $\textbf{A}_{ij}$, the proposed LAConv can be expressed as follows:
\begin{equation}\label{eq:4}
\mathbf{O}_{ij} =  \mathbf{A}_{ij}\otimes \mathbf{\widetilde{K}}_{ij} ,
\end{equation}
In particular, the generation of $\mathbf{\widetilde{K}}_{ij}$ contains following three steps as shown in Fig.~\ref{Fig:laconv}. Firstly, $\mathbf{A}_{ij}$ will be sent to the convolutional layer with the ReLU activation to yield its shallow feature. Secondly, the shallow feature will be sent to the fully connected (FC) layers with ReLU and sigmoid activations, then a weight $\mathbf{\widetilde{W}}_{ij} \in \mathbb{R}^{1\times k^{2}}$ that can perceive the potential relationship between the central pixel $\textbf{I}_{ij}$ and its neighbors is learned. Thirdly, the $\mathbf{\widetilde{W}}_{ij} \in \mathbb{R}^{1\times k^{2}}$ is reshaped to $\mathbf{W}_{ij} \in \mathbb{R}^{k \times k}$ used as the scaling factor for every kernel in $\textbf{K}$, the scaled kernel is represented as $\mathbf{\widetilde{K}}_{ij}$, whose generation process is expressed as follows:
\begin{equation}\label{eq:5}
\mathbf{\widetilde{K}}_{ij} =\mathbf{W^{D}}_{ij}\odot \mathbf{K} ,
\end{equation}
where $\odot$ represents the pixel-wise multiplication. $\mathbf{W^{D}}_{ij}$ is duplicated $\mathbf{W}_{ij}$. More details please refer to the top part in Fig.~\ref{Fig:laconv}.

\subsection{Dynamic Bias}

\begin{figure*}[t]
		\begin{center}
		{\includegraphics[width=1\linewidth]{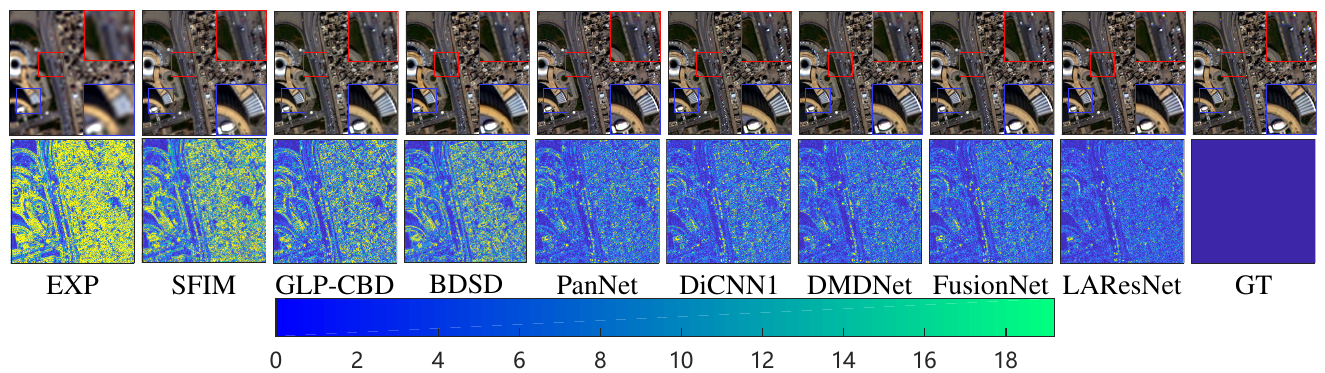}}
	\end{center}
		\vspace{3pt}
		\caption{{Qualitative comparison for pansharpening on reduced resolution Rio-dataset (source: WV3). The first row presents the RGB visualization, while the second row displays the corresponding absolute error maps (AEMs).}}\label{wv3visual}
\end{figure*}

\begin{figure*}[t]
			\begin{center}
		{\includegraphics[width=1\linewidth]{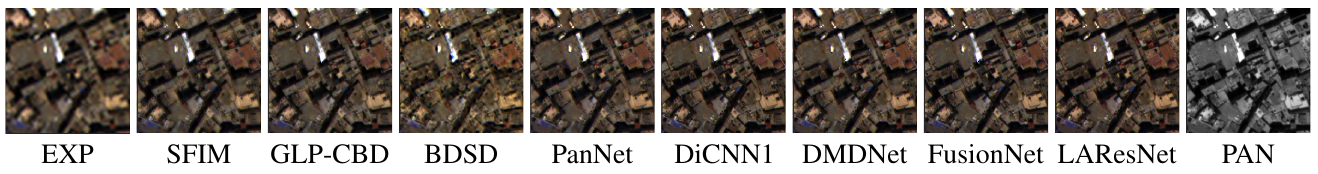}}
	\end{center}
		
		\vspace{3pt}	
		\caption{Qualitative comparison for pansharpening on full resolution WV3 dataset.}\label{Fig:wv3org}
\end{figure*}

In this section, we design DYB for our LAConv. Different from the conventional bias, the DYB is generated from the global inputting feature, which is denoted as $\mathbf{I} \in \mathbb{R}^{H \times W \times C_{in}}$. The operation process of LAConv with the conventional bias can be expressed as follows:
\begin{equation}\label{eq:7}
\mathbf{O}_{ij} =  \widetilde{\textbf{O}}_{ij} + \mathbf{D} ,
\end{equation}
where $\mathbf{D} \in \mathbb{R}^{1 \times C_{out}}$ is defined as the DYB, which is generated by the following two steps.  Firstly, the inputting feature $\mathbf{I}$ will pass through the global average pooling layer (GAP) to obtain $\mathbf{\widetilde{I}} \in \mathbb{R}^{1 \times C_{in}}$. Secondly, $\mathbf{\widetilde{I}}$ will be sent to the FC layers with ReLU activations, and the output is $\mathbf{D}$. More details can be referred to the bottom part in Fig.~\ref{Fig:laconv}.


\subsection{Proposed LAResNet}
In this work, we mainly perform two different image fusion tasks, namely pansharpening and HISR, whose source data are different. For convenience of explanation, we will uniformly denote the LR-MSI in pansharpening and LR-HSI in HISR as $``\textbf{LR}"$, and unify the HR-PANI in pansharpening and HR-MSI in HISR as $``\textbf{HR}"$. We aim to develop a simple and efficient image fusion network that takes an upsampled $``\textbf{LR}"$ (denoted as $\widetilde{\textbf{LR}}$) and an $``\textbf{HR}"$ as input and output a fused image $``\textbf{SR}"$. Fig.~\ref{Fig:laresnet} shows a detailed architecture of the proposed LAResNet.

Before introducing the architecture, it is necessary to illustrate the important components of the network, called LAResBlock. In fact, LAResBlock is exactly the same as the original ResBlock~\cite{resnet}, except that the standard convolution in ResBlock is substituted by the proposed LAConv. In what follows, we will introduce overall architecture of LAResNet. As shown in Fig.~\ref{Fig:laresnet}, the proposed network has three stages. The first stage contains a LAConv layer and an activation layer, followed by several stacked LAResBlocks. And the last stage is also a LAConv layer. Specifically, the $\textbf{HR}$ and the $\widetilde{\textbf{LR}}$ are first concatenated together to obtain a feature map $\textbf{M}$, which contains the information of the two input images. After that, $\textbf{M}$ will pass through the network stage after stage. Finally, the output of the network will be added with $\widetilde{\textbf{LR}}$ as the final $\textbf{SR}$ image. The whole processing can be expressed by the following equation:
\begin{equation}\label{eq:1}
\mathbf{SR}= \mathbf{\widetilde{LR}} +  \mathbf{\mathcal F}_{\theta }(\mathbf{\widetilde{LR}};\mathbf{HR}) ,
\end{equation}
where $\mathbf{\mathcal F}_{\theta }(\cdot)$ represents the mapping functional with its parameters $\theta$, which is updated to minimize the distance between the $\mathbf{SR}$ and the ground-truth ($\mathbf{GT}$) image. Here we also just chose the simplest mean square error (MSE) loss, the loss function can be expressed as follows:
\begin{equation}\label{eq:2}
\mathcal L(\theta ) = \frac{1}{N}\sum_{i=1}^{N}\left \| \mathcal F_{\theta}({\mathbf{\widetilde{LR}}}^{(i)};\textbf{HR}^{(i)}) + {\mathbf{\widetilde{LR}}}^{(i)} -  \textbf{GT}^{(i)} \right \|_{F}^{2} ,
\end{equation}
where $N$ is the number of training examples, and $\left \|\cdot \right \|_{F}$ is Frobenius norm.


\begin{figure*}[t]
				\begin{center}
		{\includegraphics[width=1\linewidth]{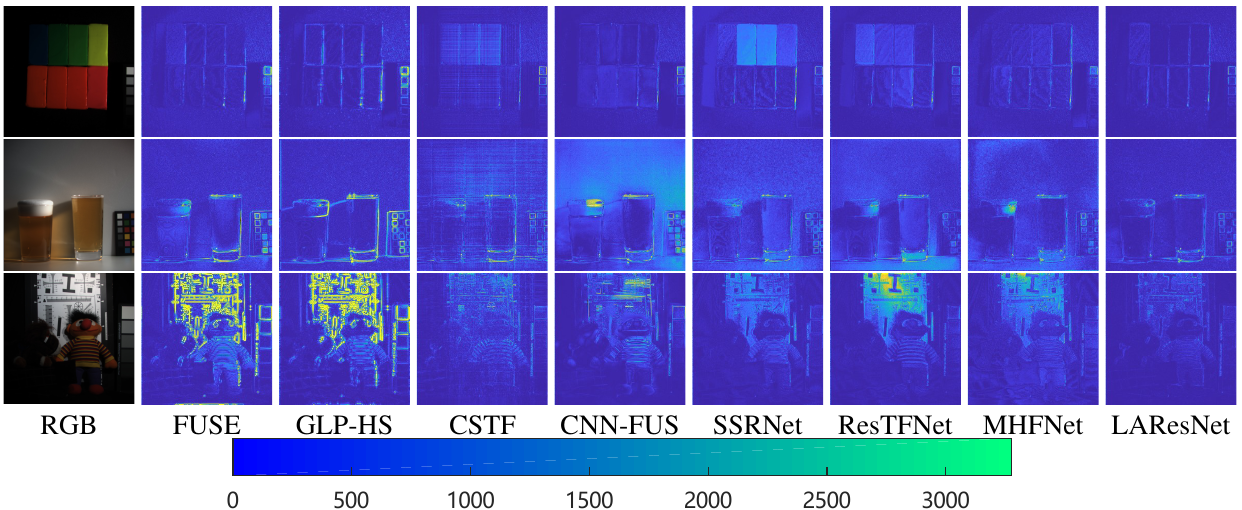}}
	\end{center}
		
		\vspace{3pt}
		\caption{AEMs comparison for HISR on CAVE dataset. }\label{Fig:cave11}
\end{figure*}

\section{Experiments}
In this section, we separately discuss the quantitative and qualitative results of the experiments for the two tasks, {\it i.e.}, pansharpening and HISR.

\begin{table}[!b]
	\scriptsize
	\setlength{\tabcolsep}{3.6pt}
	\renewcommand\arraystretch{1.15}
	\centering
	\caption{{Average quantitative comparisons on 1258 reduced resolution WV3 examples.} 
		\label{Table:WV3}}
	
	\begin{tabular}{l|cccc}
		\hline
		\textbf{Method}&\emph{SAM}&\emph{ERGAS} & \emph{SCC}& \emph{Q8}\\
		\hline
		\textbf{SFIM~\cite{sfim}}    &5.452 $\pm$ 1.903 & 4.690 $\pm$ 6.574 & 0.866 $\pm$ 0.067 & 0.798 $\pm$ 0.122 \\
		\textbf{GLP-CBD~\cite{glp_cbd}}     &5.286 $\pm$ 1.958 & 4.163 $\pm$ 1.775 & 0.890 $\pm$ 0.070 & 0.854 $\pm$ 0.114 \\
		\textbf{BDSD~\cite{bdsd}}    &7.000 $\pm$ 2.853 & 5.167 $\pm$ 2.248 & 0.871 $\pm$ 0.080 & 0.813 $\pm$ 0.123 \\
		\textbf{PanNet~\cite{pannet}}  &4.092 $\pm$ 1.273 & 2.952 $\pm$ 0.978 & 0.949 $\pm$ 0.046 & 0.894 $\pm$ 0.117 \\
		\textbf{DiCNN1~\cite{dicnn}}  &3.981 $\pm$ 1.318 & 2.737 $\pm$ 1.016 & 0.952 $\pm$ 0.047 & 0.910 $\pm$ 0.112 \\
		\textbf{DMDNet~\cite{dmdnet}}  &3.971 $\pm$ 1.248 & 2.857 $\pm$ 0.966 & 0.953 $\pm$ 0.045 & 0.913 $\pm$ 0.115 \\	
		\textbf{FusionNet~\cite{fusionnet}} &3.744 $\pm$ 1.226 & 2.568 $\pm$ 0.944 & 0.958 $\pm$ 0.045 & 0.914 $\pm$ 0.112 \\
		\textbf{LAResNet}  &\bf{3.473 $\pm$ 1.197} & \bf{2.338 $\pm$ 0.911} &\bf{0.965 $\pm$ 0.043} & \bf{0.923 $\pm$ 0.114}\\
		\hline
		\textbf{Ideal value}&\bf{0}&\bf{0}&\bf{1}&\bf{1}\\ 
		\hline
	\end{tabular}
\end{table}

\subsection{Results for Pansharpening}
In this section, we will first introduce the training implementation, then, datasets and evaluation indicators will be described, and finally, our pansharpening results will be presented.
\subsubsection{Training Details and Parameters}
\label{sec:para}
The models are implemented with PyTorch. For the parameters of LAResNet, the number of the LAResBlock is set to 5 (i.e., $B = 5$), while the channels of the LAConv and the kernel size are 32 and $k \times k$ (i.e., $k = 3$), respectively. Besides, we set 1000 epochs for the network training, while the learning rate $1 \times 10^{-3}$ in the first 500 epochs and $1 \times 10^{-4}$ in the last 500 epochs. The FC layers used in the LAConv consist of two dense layers with $k^{2}$ neurons, and the FC layers in the DYB consist of two dense layers with $C_{out}$ neurons. Adam optimizer is used for training with the batch size 32 while $\beta _{1}$ and $\beta _{2}$ are set to 0.9 and 0.999, respectively.

\begin{table}[!b]
	\scriptsize
	\setlength{\tabcolsep}{3.6pt}
	\renewcommand\arraystretch{1.15}
	\centering
	\caption{{Average quantitative comparisons on 48 reduced resolution QB examples.} 
		\label{Table:QB-48}}
	\begin{tabular}{l|cccc}
		\hline
		\textbf{Method}&\emph{SAM}&\emph{ERGAS} & \emph{SCC}& \emph{Q4}\\
		\hline
		\textbf{SFIM~\cite{sfim}}    &7.718 $\pm$ 1.872 & 8.778 $\pm$ 2.380 & 0.832 $\pm$ 0.105 & 0.767 $\pm$ 0.119 \\
		\textbf{GLP-CBD~\cite{glp_cbd}}     &7.398 $\pm$ 1.783 & 7.297 $\pm$ 0.932 & 0.854 $\pm$ 0.064 & 0.819 $\pm$ 0.128 \\
		\textbf{BDSD~\cite{bdsd}}    &7.671 $\pm$ 1.911 & 7.466 $\pm$ 0.991 & 0.851 $\pm$ 0.062 & 0.813 $\pm$ 0.136 \\
		\textbf{PanNet~\cite{pannet}}  &5.314 $\pm$ 1.018 & 5.162 $\pm$ 0.681 & 0.930 $\pm$ 0.059 & 0.883 $\pm$ 0.140 \\
		\textbf{DiCNN1~\cite{dicnn}}  &5.307 $\pm$ 0.996 & 5.231 $\pm$ 0.541 & 0.922 $\pm$ 0.051 & 0.882 $\pm$ 0.143 \\
		\textbf{DMDNet~\cite{dmdnet}}  &5.120 $\pm$ 0.940 & 4.738 $\pm$ 0.649 & 0.935 $\pm$ 0.065 & 0.891 $\pm$ 0.146 \\	
		\textbf{FusionNet~\cite{fusionnet}}  &4.540 $\pm$ 0.779 & 4.051 $\pm$ 0.267 & 0.955 $\pm$ 0.046 & 0.910 $\pm$ 0.136 \\	
		\textbf{LAResNet}  &\bf{4.378 $\pm$ 0.727} & \bf{3.740 $\pm$ 0.298} &\bf{0.959 $\pm$ 0.047} & \bf{0.916 $\pm$ 0.134}\\
		\hline
		\textbf{Ideal value}&\bf{0}&\bf{0}&\bf{1}&\bf{1}\\ 
		\hline
	\end{tabular}
\end{table}

\subsubsection{Datasets and Evaluation Metrics}
To benchmark the effectiveness of LAResNet for pansharpening, we adopt a wide range of datasets including 8-band datasets captured by WorldView-3 (WV3), 4-band datasets captured by GaoFen-2 (GF2) and QuickBird (QB) satellites. As the ground truth (GT) images are not available, Wald’s protocol~\cite{exp4} is performed to ensure the baseline image generation. All the source data can be download from the public website. For WV3-data, we obtain 12580 HR-PANI/LR-MSI/GT image pairs ($70\%$/$20\%$/$10\%$ as training/validation/testing dataset) with the size 64$\times$64$\times$1, 16$\times$16$\times$8, and 64$\times$64$\times$8, respectively; For GF2 data, we use 10000 PAN/MS/GT image pairs ($70\%$/$20\%$/$10\%$ as training/validation/testing dataset) with the size 64$\times$64$\times$1, 16$\times$16$\times$4, and 64$\times$64$\times$4, respectively; For QB data, 20000 PAN/MS/GT image pairs ($70\%$/$20\%$/$10\%$ as training/validation/testing dataset) with the size 64$\times$64$\times$1, 16$\times$16$\times$4, and 64$\times$64$\times$4 were adopted.

The quality evaluation is conducted both at reduced and full resolutions. For reduced resolution test, the spectral angle mapper (SAM)~\cite{sam}, the relative dimensionless global error in synthesis (ERGAS)~\cite{ergas}, the spatial correlation coefficient (SCC)~\cite{SCC}, and quality index for 4-band images (Q4)~\cite{q2n} and 8-band images (Q8)~\cite{q2n} are used to assess the quality of the results. In addition, to assess the performance of all involved methods on full resolutions, the QNR~\cite{QNR}, the $D_{\lambda}$~\cite{QNR}, and the $D_{s}$~\cite{QNR} indexes are applied.

\subsubsection{Comparison with State-of-the-art}

\begin{figure*}[t]
				\begin{center}
		{\includegraphics[width=1\linewidth]{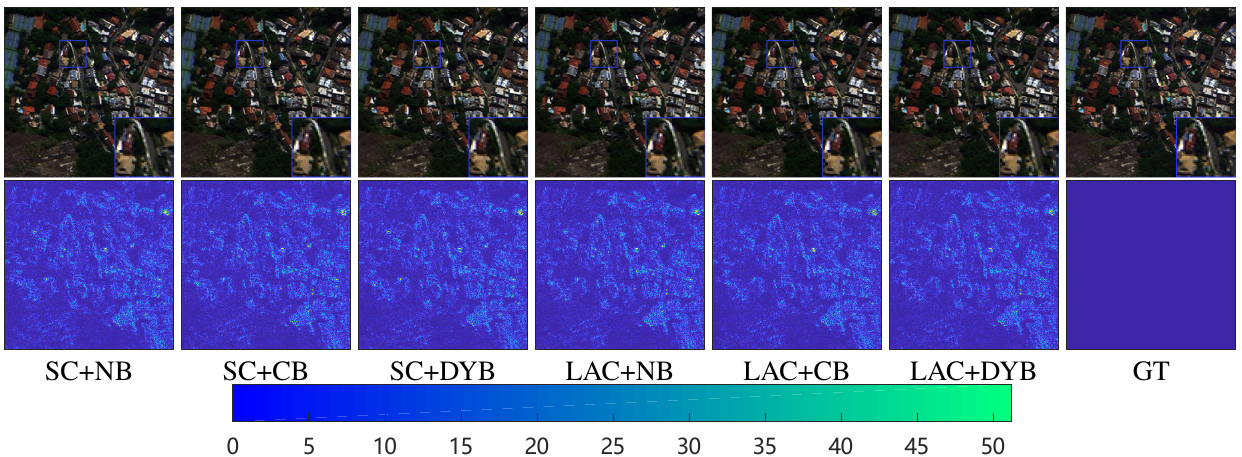}}
	\end{center}

		\vspace{3pt}
		\caption{Qualitative comparison for pansharpening of ablation study on reduced WV3 dataset. }\label{Fig:ablation}
\end{figure*}

\begin{table}[!t]
	\scriptsize
	\setlength{\tabcolsep}{3.6pt}
	\renewcommand\arraystretch{1.15}
	\centering
	\caption{{Average quantitative comparisons on 81 reduced resolution GF2 examples.} 
		\label{Table:GF-81}}
	\begin{tabular}{l|cccc}
		\hline
		\textbf{Method}&\emph{SAM}&\emph{ERGAS} & \emph{SCC}& \emph{Q4}\\
		\hline
		\textbf{SFIM~\cite{sfim}}    &2.297 $\pm$ 0.638 & 2.189 $\pm$ 0.695 & 0.861 $\pm$ 0.054 & 0.865 $\pm$ 0.040 \\
		\textbf{GLP-CBD~\cite{glp_cbd}}     &2.274 $\pm$ 0.733 & 2.046 $\pm$ 0.620 & 0.873 $\pm$ 0.053 & 0.877 $\pm$ 0.041 \\
		\textbf{BDSD~\cite{bdsd}}    &2.307 $\pm$ 0.670 & 2.070 $\pm$ 0.610 & 0.877 $\pm$ 0.052 & 0.876 $\pm$ 0.042 \\
		\textbf{PanNet~\cite{pannet}}  &1.400 $\pm$ 0.326 & 1.224 $\pm$ 0.283 & 0.956 $\pm$ 0.012 & 0.947 $\pm$ 0.022 \\
		\textbf{DiCNN1~\cite{dicnn}}  &1.495 $\pm$ 0.381 & 1.320 $\pm$ 0.354 & 0.946 $\pm$ 0.022 & 0.945 $\pm$ 0.021 \\
		\textbf{DMDNet~\cite{dmdnet}}  &1.297 $\pm$ 0.316 & 1.128 $\pm$ 0.267 & 0.964 $\pm$ 0.010 & 0.953 $\pm$ 0.022 \\	
		\textbf{FusionNet~\cite{fusionnet}}  &1.180 $\pm$ 0.271 & 1.002 $\pm$ 0.227 & 0.971 $\pm$ 0.007 & 0.963 $\pm$ 0.017 \\	
		\textbf{LAResNet}  &\bf{1.085 $\pm$ 0.238} & \bf{0.912 $\pm$ 0.206} &\bf{0.977 $\pm$ 0.006} & \bf{0.970 $\pm$ 0.016}\\
		\hline
		\textbf{Ideal value}&\bf{0}&\bf{0}&\bf{1}&\bf{1}\\ 
		\hline
	\end{tabular}
\end{table}

In this section, we will show the comparison of the results on various datasets obtained by our LAResNet and several competitive methods (including traditional methods and DL-based methods), which were introduced in Sec.~\ref{pansharpening}.

{\bf Evaluation on 8-band reduced resolution dataset.} We compare the proposed method with recent state-of-the-art pansharpening methods on the quantitative performance on 1258 WV3 testing datasets. The results of compared methods and LAResNet are reported in Tab.~\ref{Table:WV3}. It can be observed that LAResNet achieves a transcendence performance. Also we compare the related approaches on the Rio-dataset (WV3), whose visual results are shown in Fig.~\ref{wv3visual}. It can be seen that our result is the closest to the GT image.

{\bf Evaluation on 8-band full resolution dataset.} We further perform a full resolution test experiment on the WV3 dataset with  50 pairs. The quantitative results are reported in Tab.~\ref{tab:WV3-51}, and the visual results are shown in Fig.~\ref{Fig:wv3org}. Again, our method also surpasses other methods both in visual comparison and quantitative indicators.

{\bf Evaluation on 4-band reduced resolution dataset.} In order to prove the wide applicability of LAResNet, we also conducted experiments on the 4-band GF2 and QB datasets. Similarly, the comparison of quantified indicators is shown in Tab.~\ref{Table:QB-48} and Tab.~\ref{Table:GF-81}, which indicates that our method can produce the best outcomes whether the GF2 or QB data.

\subsection{Results for HISR}

\subsubsection{Training Details and Parameters}

We conduct 550 epochs training under the Pytorch framework, and the learning rate is fixed as $1 \times 10^{-3}$ during the training process. For the parameters of LAResNet, the number of the LAResBlock is set to 3 (i.e., $B = 3$), while the channels of the LAConv is set to 64. The rest of the settings and parameters are the same as that in Sec.~\ref{sec:para}

\begin{table}[!t]
	\scriptsize
	\setlength{\tabcolsep}{6.7pt}
	\renewcommand\arraystretch{1.15}
	\centering
	\caption{\small{Average quantitative comparisons on 50 full resolution WV3 examples.} 
		\label{tab:WV3-51}}
	\begin{tabular}{l|ccc}
		\hline
		\textbf{Method}&\emph{QNR}&\emph{$D_{\lambda}$} & \emph{$D_{s}$}\\
		\hline
		\textbf{SFIM~\cite{sfim}}    &0.9282 $\pm$ 0.0512 & 0.0254 $\pm$ 0.0287 & 0.0485 $\pm$ 0.0283 \\
		\textbf{GLP-CBD~\cite{glp_cbd}} &0.9113 $\pm$ 0.0671 & 0.0331 $\pm$ 0.0338 & 0.0590 $\pm$ 0.0432 \\
		\textbf{BDSD~\cite{bdsd}}    &0.9300 $\pm$ 0.0491 & 0.0177 $\pm$ 0.0130 & 0.0537 $\pm$ 0.0404 \\
		\textbf{PanNet~\cite{pannet}}  &0.9521 $\pm$ 0.0219 & 0.0260 $\pm$ 0.0114 &0.0226 $\pm$ 0.0123 \\
		\textbf{DiCNN1~\cite{dicnn}}  &0.9436 $\pm$ 0.0458 & 0.0185 $\pm$ 0.0210 & 0.0392 $\pm$ 0.0299 \\
		\textbf{DMDNet~\cite{dmdnet}}  &0.9554 $\pm$ 0.0200 & 0.0215 $\pm$ 0.0099 & 0.0237 $\pm$ 0.0118 \\
		\textbf{FusionNet~\cite{fusionnet}} &0.9556 $\pm$ 0.0316 & 0.0198 $\pm$ 0.0168 & 0.0254 $\pm$ 0.0183 \\	
		\textbf{LAResNet}  &\bf{0.9637 $\pm$ 0.0119} & \bf{0.0147 $\pm$ 0.0077} & \bf{0.0220 $\pm$ 0.0064} \\
		\hline
		\textbf{Ideal value}&\bf{1}&\bf{0}&\bf{0}\\ 
		\hline
	\end{tabular}
\end{table}

\begin{table}[!b]
	\scriptsize
	\setlength{\tabcolsep}{4.8pt}
	\renewcommand\arraystretch{1.15}
	\centering
	\caption{{Average quantitative comparisons on 11 CAVE examples.} 
		\label{Table:cave11}}
	\begin{tabular}{l|cccc}
		\hline
		\textbf{Method}&\emph{PSNR}&\emph{SAM} & \emph{ERGAS}& \emph{SSIM}\\
		\hline
		\textbf{FUSE~\cite{FUSE}}    &39.72 $\pm$ 3.52& 5.83 $\pm$ 2.02& 4.18 $\pm$ 3.08& 0.975 $\pm$ 0.018\\
		\textbf{GLP-HS~\cite{GLP-HS}}    &37.81 $\pm$ 3.06& 5.36 $\pm$ 1.78& 4.66 $\pm$ 2.71& 0.972 $\pm$ 0.015\\
		\textbf{CSTF~\cite{CSTF}}    &42.14 $\pm$ 3.04& 9.92 $\pm$ 4.11& 3.08 $\pm$ 1.56& 0.964 $\pm$ 0.027\\
		\textbf{CNN-FUS~\cite{CNN-FUS}}    &42.66 $\pm$ 3.46& 6.44 $\pm$ 2.31& 2.95 $\pm$ 2.24& 0.982 $\pm$ 0.007\\
		\textbf{SSRNet~\cite{SSRNET}}    &45.28 $\pm$ 3.13& 4.72 $\pm$ 1.76& 2.06 $\pm$ 1.30& 0.990 $\pm$ 0.004\\
		\textbf{ResTFNet~\cite{ResTFnet}}    &45.35 $\pm$ 3.68& 3.76 $\pm$ 1.31& 1.98 $\pm$ 1.62& 0.993 $\pm$ 0.003\\
		\textbf{MHFNet~\cite{MHFNet}}    &46.32 $\pm$ 2.76& 4.33 $\pm$ 1.48& 1.74 $\pm$ 1.44& 0.992 $\pm$ 0.006\\
		\textbf{LAResNet}    &\bf{47.68 $\pm$ 3.37} & \bf{3.07 $\pm$ 0.97} & \bf{1.49 $\pm$ 0.96} & \bf{0.995 $\pm$ 0.002}\\
		\hline
		\textbf{Ideal value}&\bf{$\infty$}&\bf{0}&\bf{0}&\bf{1}\\ 
		\hline
	\end{tabular}
\end{table}

\begin{figure*}[!htp]
	\begin{center}
		{\includegraphics[width=1\linewidth]{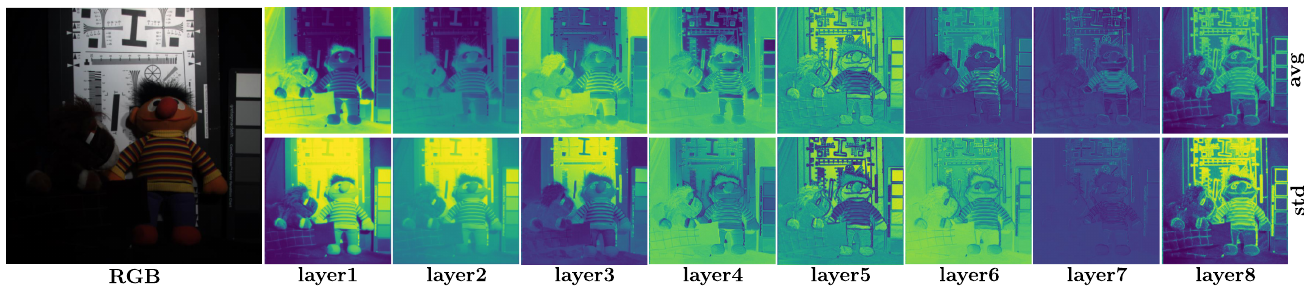}}
	\end{center}
	\caption{{Average (avg) $\mathbf{W}_{ij}$ and its standard deviation (std) on each layer  by the proposed LAConv. Note that since the values in $\mathbf{W}_{ij}$ vary dramatically on different layer, thus we do not show colorbars in this figure for better presentation.}}\label{Fig:caverlt}
\end{figure*}
\begin{table}[!b]
	\scriptsize
	\setlength{\tabcolsep}{5.5pt}
	\renewcommand\arraystretch{1.15}
	\centering
	\caption{{Quantitative comparisons of ablation study on Tripoli-dataset (WV3).} 
		\label{Table:abli}}
	\begin{tabular}{l|cccc}
		\hline
		\textbf{Method}&\emph{SAM}&\emph{ERGAS} & \emph{SCC}& \emph{Q8}\\
		\hline
		\textbf{SC + NB}    &4.2564 & 3.1026 & 0.9628 & 0.9511 \\
		\textbf{SC + CB}    &4.3483 & 3.1302 & 0.9614 & 0.9511 \\
		\textbf{SC + DYB}    &4.2267 & 3.0575 & 0.9637 & 0.9524 \\
		\textbf{LAC + NB}    &4.0354 & 2.9495 & 0.9676 & 0.9571 \\
		\textbf{LAC + CB}    &4.0264 & 2.9129 & 0.9684 & 0.9568 \\
		\textbf{LAC + DYB}    &\bf{3.9740} & \bf{2.9010} & \bf{0.9692} & \bf{0.9584} \\
		\hline
		\textbf{Ideal value}&\bf{0}&\bf{0}&\bf{1}&\bf{1}\\ 
		\hline
	\end{tabular}
\end{table}

\begin{table}[t]
	\renewcommand\arraystretch{1.5}
	\caption{\small{Report on number of parameters (NoPs). First two lines is the pansharpening experiment on WV3 dataset, and the last two lines are HISR experiment. } \label{tab:para}}
	\footnotesize
	\begin{center}
		\resizebox{0.48\textwidth}{!}{  
			\begin{tabular}{l|cccccc}
				\hline
				\textbf{Method}& \textbf{PNN}&	\textbf{DiCNN1} &\textbf{PanNet} &\textbf{DMDNet}	 &\textbf{FusionNet}&\textbf{LAResNet} \\
				\hline
				\textbf{NoPs}	& $3.1\times 10^5$  &	$1.8\times 10^5$  & $2.5\times 10^5$  &   $3.2\times 10^5$ & $2.3\times 10^5$&  $1.5\times 10^5$  \\
				\hline
		\end{tabular}}
	\end{center}
\end{table}

\subsubsection{Datasets and Evaluation Metrics}
In this work, we mainly adopt the CAVE dataset~\cite{cavedata} for training the network. We have simulated a total of 3920 HR-MSI/LR-HSI/GT image pairs ($80\%$/$20\%$ as training/testing dataset) with the size 64$\times$64$\times$3, 16$\times$16$\times$31, and 64$\times$64$\times$31, respectively. The process of data generation contains the following three steps: 1) Crop 3920 overlapping patches from the original CAVE dataset as GT, also called HR-HSI; 2) Apply a Gaussian blur with the kernel size of 3$\times$3 and standard deviation of 0.5 to HR-HSI patches, and then the blurred patches are downsampled to generate LR-HSI patches; 3) Use the spectral response function of Nikon D700 camera~\cite{nikon1,CNN-FUS,MHFNet,nikon2} to generate RGB patches. Besides, to evaluate the performance of HISR, we adopt the following indicators, SAM, ERGAS, the peak signal-to-noise ratio (PSNR) and the structure similarity (SSIM)~\cite{SSIM}.

\subsubsection{Comparison with State-of-the-art}

In this section, we will report the comparison of the results on CAVE datasets produced by our LAResNet and several advanced methods (including traditional methods and DL-based methods), which were introduced in Sec.~\ref{hisr}. Quantitative and qualitative evaluation results of these approaches are summarized in Tab.~\ref{Table:cave11} and Fig.~\ref{Fig:cave11}. As can be observed, our method exceeds the state-of-the-art methods significantly from the perspective of visual effects. Besides the significant improvement of pansharpening performance, our proposed LAResNet also has a favorable performance for HISR. We believe that LAConv can also achieve satisfactory results for more super-resolution tasks.

\subsection{Ablation Study}
In order to verify the effectiveness of LAConv and DYB, we perform six groups of ablation on WV3 dataset. In the case of the same number of residual blocks and channels, six modes are performed. The specific settings are standard convolution + no bias (SC+NB), standard convolution + conventional bias (SC+CB), standard convolution + dynamic bias (SC+DYB), LAConv + no bias (LAC+NB), LAConv + conventional bias (LAC+CB), LAConv + dynamic bias (LAC+DYB, i.e., the proposed).  The experimental results are shown in Fig.~\ref{Fig:ablation} and Tab.~\ref{Table:abli}. It is clear that the network with LAConv works better than the network with standard Conv. And the comparison between conventional bias and no bias indicates that the conventional bias is worse than the no bias network, which demonstrates the conventional bias is not suitable for this image fusion task. The network with dynamic bias, on the other hand, is significantly more comparable than the network with conventional bias and no bias under the same settings.

\subsection{More Discussions}

{\bf Exploration on LAConv} To better illustrate the effectiveness of LAConv in the fusion process, we present the average and variance of the weights ($\textbf{W}_{ij}$) by LAConv for each convolution layer in Fig.~\ref{Fig:caverlt}. Through analysis, it can be seen that $\textbf{W}_{ij}$ generated by LAConv in different local areas are different, more specifically, LAConv mainly focuses on the overall information of objects in the shallow layers, and then focuses on the local high-frequency features such as edges with the layers going deeper.

{\bf Report on the number of parameters} The number of parameters (NoPs) of all the compared CNNs are presented in Tab.~\ref{tab:para}. It can be seen that the amount of parameters of LAResNet is the least, whereas the best results are achieved.

\section{Conclusion}
We have presented a novel local adaptive convolution (LAConv) and dynamic bias (DYB) for image fusion. Leveraging on the locally adaptive dynamic convolution kernel, LAConv has powerful local focusing and feature representation capabilities. We further propose a simple residual structure network equiped with LAConv and DYB called LAResNet for two image fusion tasks. Experiments prove that our method achieves the state-of-the-art results both in pansharpening and HISR. The adaptive local focusing mechanism and translation-invariance property of the LAConv guarantee its huge potential for other pixel-wise vision tasks, such as single image super-resolution or image classification.


\section{Acknowledge}
This work is supported by NSFC (61702083), Key Projects of Applied Basic Research in Sichuan Province (Grant No. 2020YJ0216), and National Key Research and Development
Program of China (Grant No. 2020YFA0714001).

\bibliographystyle{unsrt}  


\bibliography{egbib}

\begin{thebibliography}{10}

\bibitem{condconv}
Brandon Yang, Gabriel Bender, Quoc~V Le, and Jiquan Ngiam.
\newblock Condconv: Conditionally parameterized convolutions for efficient
  inference.
\newblock {\em Advances in Neural Information Processing Systems},
  32:1307--1318, 2019.

\bibitem{dynet}
Yikang Zhang, Jian Zhang, Qiang Wang, and Zhao Zhong.
\newblock Dynet: Dynamic convolution for accelerating convolutional neural
  networks.
\newblock {\em arXiv preprint arXiv:2004.10694}, 2020.

\bibitem{drconv}
Jin Chen, Xijun Wang, Zichao Guo, Xiangyu Zhang, and Jian Sun.
\newblock Dynamic region-aware convolution.
\newblock {\em arXiv preprint arXiv:2003.12243}, 2020.

\bibitem{sfim}
{J. Liu}.
\newblock Smoothing filter-based intensity modulation: A spectral preserve
  image fusion technique for improving spatial details.
\newblock {\em International Journal of Remote Sensing}, 21(18):3461--3472,
  2000.

\bibitem{glp}
Bruno Aiazzi, Luciano Alparone, Stefano Baronti, and Andrea Garzelli.
\newblock Context-driven fusion of high spatial and spectral resolution images
  based on oversampled multiresolution analysis.
\newblock {\em IEEE Transactions on Geoscience and Remote Sensing},
  40(10):2300--2312, 2002.

\bibitem{mtf}
B~Aiazzi, L~Alparone, S~Baronti, A~Garzelli, and M~Selva.
\newblock Mtf-tailored multiscale fusion of high-resolution ms and pan imagery.
\newblock {\em Photogrammetric Engineering \& Remote Sensing}, 72(5):591--596,
  2006.

\bibitem{glp_cbd}
Luciano Alparone, Lucien Wald, Jocelyn Chanussot, Claire Thomas, Paolo Gamba,
  and Lori~Mann Bruce.
\newblock Comparison of pansharpening algorithms: Outcome of the 2006 grs-s
  data-fusion contest.
\newblock {\em IEEE Transactions on Geoscience and Remote Sensing},
  45(10):3012--3021, 2007.

\bibitem{bdsd}
Andrea Garzelli, Filippo Nencini, and Luca Capobianco.
\newblock Optimal mmse pan sharpening of very high resolution multispectral
  images.
\newblock {\em IEEE Transactions on Geoscience and Remote Sensing},
  46(1):228--236, 2007.

\bibitem{pannet}
Junfeng Yang, Xueyang Fu, Yuwen Hu, Yue Huang, Xinghao Ding, and John Paisley.
\newblock Pannet: A deep network architecture for pan-sharpening.
\newblock In {\em Proceedings of the IEEE International Conference on Computer
  Vision}, pages 5449--5457, 2017.

\bibitem{dicnn}
Lin He, Yizhou Rao, Jun Li, Jocelyn Chanussot, Antonio Plaza, Jiawei Zhu, and
  Bo~Li.
\newblock Pansharpening via detail injection based convolutional neural
  networks.
\newblock {\em IEEE Journal of Selected Topics in Applied Earth Observations
  and Remote Sensing}, 12(4):1188--1204, 2019.

\bibitem{dmdnet}
Xueyang Fu, Wu~Wang, Yue Huang, Xinghao Ding, and John Paisley.
\newblock Deep multiscale detail networks for multiband spectral image
  sharpening.
\newblock {\em IEEE Transactions on Neural Networks and Learning Systems},
  2020.

\bibitem{fusionnet}
Liang-Jian Deng, Gemine Vivone, Cheng Jin, and Jocelyn Chanussot.
\newblock Detail injection-based deep convolutional neural networks for
  pansharpening.
\newblock {\em IEEE Transactions on Geoscience and Remote Sensing}, 2020.

\bibitem{FUSE}
Qi~Wei, Nicolas Dobigeon, and Jean-Yves Tourneret.
\newblock Fast fusion of multi-band images based on solving a {Sylvester}
  equation.
\newblock {\em IEEE Transactions on Image Processing}, 24(11):4109--4121, 2015.

\bibitem{CSTF}
Shutao Li, Renwei Dian, Leyuan Fang, and Jose~M Bioucasdias.
\newblock Fusing hyperspectral and multispectral images via coupled sparse
  tensor factorization.
\newblock {\em IEEE Transactions on Image Processing}, 27(8):4118--4130, 2018.

\bibitem{CNN-FUS}
Renwei {Dian}, Shutao {Li}, and Xudong {Kang}.
\newblock Regularizing hyperspectral and multispectral image fusion by cnn
  denoiser.
\newblock {\em IEEE Transactions on Neural Networks and Learning Systems},
  pages 1--12, 2020.

\bibitem{SSRNET}
Xueting {Zhang}, Wei {Huang}, Qi~{Wang}, and Xuelong {Li}.
\newblock Ssr-net: Spatial-spectral reconstruction network for hyperspectral
  and multispectral image fusion.
\newblock {\em IEEE Transactions on Geoscience and Remote Sensing}, pages
  1--13, 2020.

\bibitem{ResTFnet}
Xiangyu Liu, Qingjie Liu, and Yunhong Wang.
\newblock Remote sensing image fusion based on two-stream fusion network.
\newblock {\em Information Fusion}, 55:1--15, 2020.

\bibitem{MHFNet}
Qi~Xie, Minghao Zhou, Qian Zhao, Zongben Xu, and Deyu Meng.
\newblock {MHF-Net}: An interpretable deep network for multispectral and
  hyperspectral image fusion.
\newblock volume~PP, pages 1--1, 08 2020.

\bibitem{resnet}
Kaiming He, Xiangyu Zhang, Shaoqing Ren, and Jian Sun.
\newblock Deep residual learning for image recognition.
\newblock In {\em Proceedings of the IEEE Conference on Computer Vision and
  Pattern Recognition}, pages 770--778, 2016.

\bibitem{exp4}
S.~Baronti B.~Aiazzi, L.~Alparone and A.~Garzelli.
\newblock Context-driven fusion of high spatial and spectral resolution images
  based on oversampled multiresolution analysis.
\newblock {\em IEEE Transactions on Geoscience and Remote Sensing},
  40(10):2300--2312, 2002.

\bibitem{sam}
Roberta~H. Yuhas, Alexander F.~H. Goetz, and Joe~W. Boardman.
\newblock Discrimination among semi-arid landscape endmembers using the
  spectral angle mapper ({s}{a}{m}) algorithm.
\newblock {\em JPL Airborne Geoscience Workshop; AVIRIS Workshop: Pasadena, CA,
  USA}, pages 147--149, 1992.

\bibitem{ergas}
Lucien Wald.
\newblock {Data fusion: definitions and architectures: Fusion of images of
  different spatial resolutions}.
\newblock {\em Presses des MINES}, 2002.

\bibitem{SCC}
J.~{Zhou}, D.~L. {Civco}, and J.~A. {Silander}.
\newblock A wavelet transform method to merge landsat tm and spot panchromatic
  data.
\newblock {\em International Journal of Remote Sensing}, 19:743--757, 1998.

\bibitem{q2n}
Andrea Garzelli and Filippo Nencini.
\newblock Hypercomplex quality assessment of multi-/hyper-spectral images.
\newblock {\em IEEE Geoscience and Remote Sensing Letters}, 6(4):662--665,
  2009.

\bibitem{QNR}
Gemine Vivone, Luciano Alparone, Jocelyn Chanussot, Mauro~Dalla Mura, Andrea
  Garzelli, Giorgio~A. Licciardi, Rocco Restaino, and Lucien Wald.
\newblock {A critical comparison among pansharpening algorithms}.
\newblock {\em IEEE Transactions on Geoscience and Remote Sensing},
  53(5):2565--2586, 2015.

\bibitem{GLP-HS}
Massimo Selva, Bruno Aiazzi, Francesco Butera, Leandro Chiarantini, and Stefano
  Baronti.
\newblock Hyper-sharpening: A first approach on {SIM-GA} data.
\newblock {\em IEEE Journal of Selected Topics in Applied Earth Observations
  and Remote Sensing}, 8(6):3008--3024, 2015.

\bibitem{cavedata}
Fumihito Yasuma, Tomoo Mitsunaga, Daisuke Iso, and Shree~K Nayar.
\newblock Generalized assorted pixel camera: postcapture control of resolution,
  dynamic range, and spectrum.
\newblock {\em IEEE Transactions on Image Processing}, 19(9):2241--2253, 2010.

\bibitem{nikon1}
Renwei Dian and Shutao Li.
\newblock Hyperspectral image super-resolution via subspace-based low tensor
  multi-rank regularization.
\newblock {\em IEEE Transactions on Image Processing}, 28(10):5135--5146, 2019.

\bibitem{nikon2}
Ting Xu, Ting-Zhu Huang, Liang-Jian Deng, Xi-Le Zhao, and Jie Huang.
\newblock Hyperspectral image super-resolution using unidirectional total
  variation with tucker decomposition.
\newblock {\em IEEE Journal of Selected Topics in Applied Earth Observations
  and Remote Sensing}, 2020.

\bibitem{SSIM}
Zhou Wang, Alan~C Bovik, Hamid~R Sheikh, and Eero~P Simoncelli.
\newblock Image quality assessment: from error visibility to structural
  similarity.
\newblock {\em IEEE Transactions on Image Processing}, 13(4):600--612, 2004.

\end{thebibliography}

\end{document}